\pdfoutput=1

\documentclass[11pt]{article}

\usepackage[]{acl}

\usepackage{times}
\usepackage{latexsym}
\usepackage{multirow}
\usepackage{booktabs}
\usepackage{graphicx}
\usepackage{amsmath}

\newcommand\Mark[1]{\textsuperscript#1}
\usepackage[T1]{fontenc}

\usepackage[utf8]{inputenc}

\usepackage{microtype}

\title{Discourse-Aware Emotion Cause Extraction in Conversations}

\author{Dexin Kong\Mark{1}, Nan Yu\Mark{1}, Yun Yuan\Mark{1}, Guohong Fu\Mark{1}\Mark{,}\Mark{2} \thanks{~Corresponding author.}, Chen Gong\Mark{1}\Mark{,}\Mark{2}\\
\Mark{1}School of Computer Science and Technology, Soochow University, China \\
\Mark{2}Institute of Artificial Intelligence, Soochow University, China \\
  \texttt{sakurakdx@gmail.com;}\\
  \texttt{\{nyu,yyuanwind\}@stu.suda.edu.cn;}\\
  \texttt{\{ghfu, gongchen18\}@suda.edu.cn}
}

\begin{document}
\maketitle

\begin{abstract}
Emotion Cause Extraction in Conversations~(ECEC) aims to extract the utterances which contain the emotional cause in conversations.
Most prior research focuses on modelling conversational contexts with sequential encoding, ignoring the informative interactions between utterances and conversational-specific features for ECEC.
In this paper, we investigate the importance of discourse structures in handling utterance interactions and conversation-specific features for ECEC.
To this end, we propose a discourse-aware model~(DAM) for this task.
Concretely, we jointly model ECEC with discourse parsing using a multi-task learning~(MTL) framework and explicitly encode discourse structures via gated graph neural network~(gated GNN),
integrating rich utterance interaction information to our model.
In addition, we use gated GNN to further enhance our ECEC model with conversation-specific features.
Results on the benchmark corpus show that DAM
outperform the state-of-the-art~(SOTA) systems in the literature. 
This suggests that the discourse structure may contain a potential link between emotional utterances and their
corresponding cause expressions.
It also verifies the effectiveness of conversational-specific features.
The codes of this paper will be available on GitHub\footnote{\url{http://github.com/}}. 
\end{abstract}

\section{Introduction}
\begin{figure}[t]
\centering
\includegraphics[width=0.95\columnwidth]{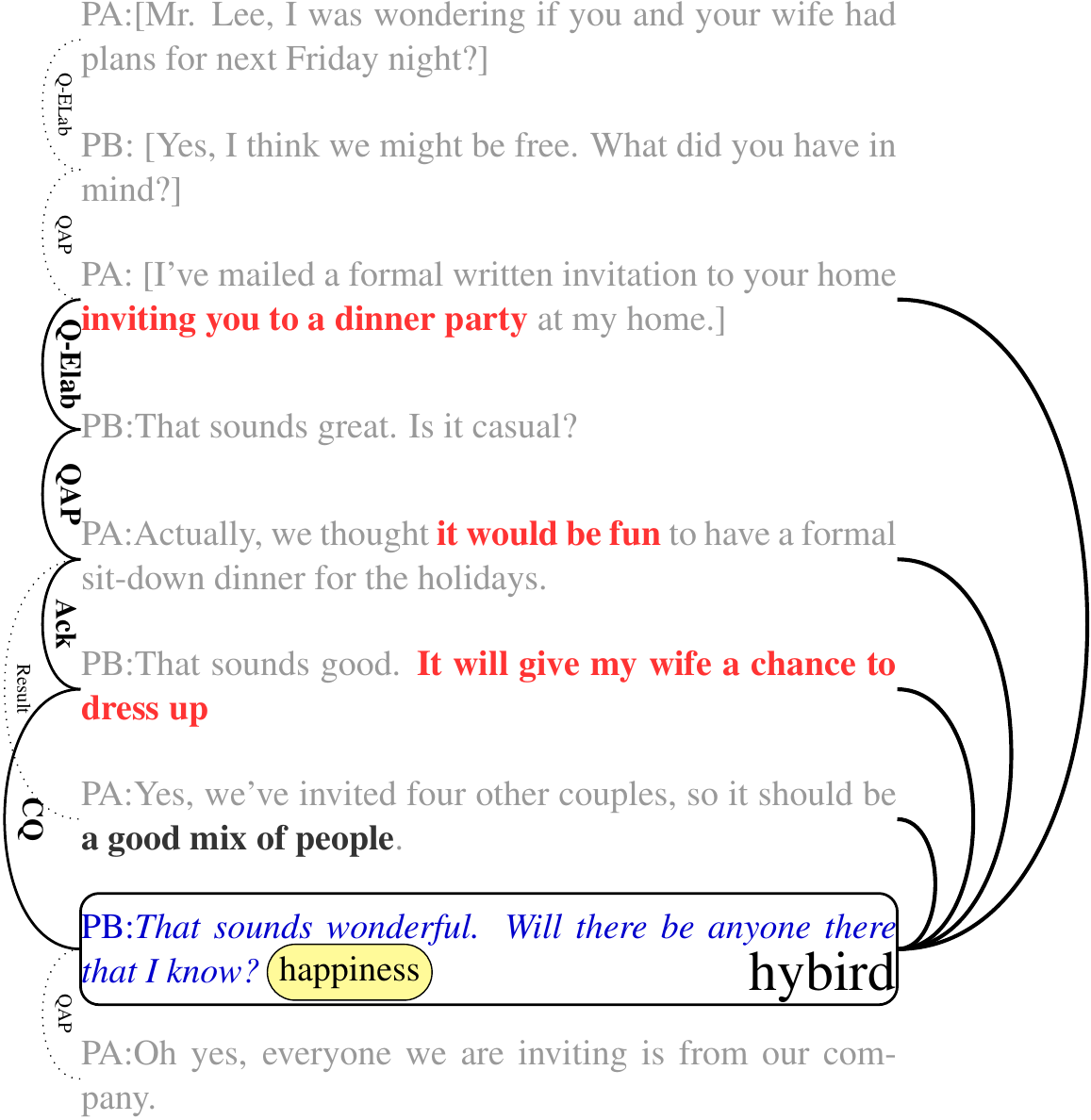}
\caption{An example of ECEC. Left arcs are predicted discourse structures. Right arcs are emotion-cause pair annotations.} 
\label{fig1}
\end{figure}

Emotion cause extraction in conversations~(ECEC) is an important task in conversation analysis.
It has received increasing attention~\cite{poria2021recognizing,multimodal_2021_wang} with the open conversational data deluge on social media platforms.
Similar to text-level emotion cause extraction~(ECE)~\cite{lee2010text}, this task aims to identify utterances that contain explanations for emotional causes in the given conversation.
The right part of Fig.~\ref{fig1} shows an example of ECEC.
The emotion of the target utterance ``\textit{That sounds wonderful. Will there be anyone there that I know?}'' is ``happiness''.
The cause utterance spans of it are ``\textit{inviting you to a dinner party}'', ``\textit{it would be fun}'', and ``\textit{It will give my wife a chance to dress up}'', 
It suggests that explicit emotion causes aforementioned could be helpful for many downstream tasks, such as opinion mining~\cite{choiIdentifyingSourcesOpinions2005, dasFindingEmotionHolder2010}.

ECE has been investigated intensively since early research~\cite{lee2010text,chen2010emotion}.
It can be treated as a classification task which requires the contexts of a document as inputs and determine whether each clause is a cause.
Recently, several neural models for ECE have been proposed~\cite{dingExperimentalStudyEffects2020,liBoundaryDetectionBERT2021a,huBidirectionalHierarchicalAttention2021}, using pre-trained language models such as BERT~\cite{devlinBertPretrainingDeep2018} and RoBERTa~\cite{liu2019roberta} to represent the plain texts.
Compared with text-level ECE, the research of emotion cause extraction in conversation is at its preliminary stage.
\citet{poria2021recognizing} propose this task for the first time, and apply neural approaches of text-level emotion cause extraction~\cite{wei2020effective,ding2020ecpe,ding2020end}.

Above traditional approaches treat conversational contexts as common plain text while using a Transformer-based encoder~\cite{liu2019roberta} to learn the conversational context representation for the ECEC task~\cite{poria2021recognizing}.
These approaches ignore the interaction information between utterances and conversational-specific features.
The problem could be serious when performing the ECEC task in long conversations.

The discourse structures in a conversation represent the interactive relationships between utterances.
Intuitively, these structures contain the potential links between emotional utterances and their corresponding cause expressions.
As shown by the solid arc on the left in Fig.~\ref{fig1}, most of the emotion-cause pairs are linked with the target utterance by discourse relations.
In addition, several findings of previous studies on ECE support our point of view.
For instance, \citet{huFSSGCNGraphConvolutional2021, dingIndependentPredictionReordered2019} believe that the interaction features between sentences could be useful information for the ECE task.


In this paper, we propose a discourse-aware model for ECEC.
Concretely, we model ECEC and discourse parsing in conversations~\cite{afantenosDiscourseParsingMultiparty2015} jointly.
It uses a shared pre-trained language model (PLM) to represent conversational contexts.
The discourse parsing task can integrate rich utterance interactions information into the shared PLM.
Besides, we use a gated graph neural network~(gated GNN)~\cite{gatedgnn2016} to explicitly encode discourse structures that generated by the discourse parser.
In addition, we follow \citet{wang2021structure}, exploiting a gated GNN to further integrate conversation-specific features such as the relative utterance distance and speakers to our model.

We conduct the experiments on the standard benchmark dataset of the ECEC task to verify our DAM model.
Experiments show that the utterance interaction features are effective for this task.
When the conversation-specific features are integrated by gated GNN,
the proposed model is able to obtain further improvements.

To sum up, we make three main contributions as follows:
\begin{itemize}
\item[$\bullet$] We propose a discourse-aware model for ECEC named DAM using multiple task learning and a gated GNN,
which is able to integrate rich utterance interaction features for ECEC.
\item[$\bullet$] We further exploit a gated GNN to capture conversation-specific features such as the relative utterance distance and speakers for ECEC.
\item[$\bullet$] We advance the performance of the SOTA models for ECEC.
\end{itemize}

We organize the rest of this paper as follows.
First, in Section~\ref{sec:related work}, we introduce the related work.
Following in Section~\ref{sec:methodlogy}, we introduce the proposed model, including the multi-task learning framework, and the gated GNN module.
Section~\ref{sec:experiments} and Section~\ref{sec:results} describes our experiments on a benchmark dataset, verifying the effectiveness of our proposed approach.
Finally, we make conclusions on Section~\ref{sec:conclusion}.

\section{Related Work}\label{sec:related work}
Prior research on emotion cause extraction can be divided into two categories according to the source text type: text-level emotion cause extraction~(ECE) and emotion cause extraction in conversation~(ECEC)~\cite{poria2021recognizing}.
For the ECE task, early research adopts linguistic rules~\cite{lee2010text,chen2010emotion,russo2011emocause} and traditional machine learning~\cite{gui2014emotion,gui2016emotion,gui2018event,xuEnsembleApproachEmotion2017}.
In recently years, several neural network models are introduced to the ECE task with different granularities, such as clause-level~\cite{diaoMultigranularityBidirectionalAttention2020,dingExperimentalStudyEffects2020,huBidirectionalHierarchicalAttention2021} and span-level~\cite{liBoundaryDetectionBERT2021a,liSpanLevelEmotionCause2021a,qianMultiTaskMRCFramework2021,turcanMultiTaskLearningAdapted2021a,liSpanlevelEmotionCause2021}.
Except for the text-level emotion cause extraction, \citet{poria2021recognizing} introduces a new conversational dataset RECCON and the Transformer-based baseline models. 

Many researchers realize that there is a connection between ECE task and emotion recognition task.
In order to make full use of the interaction between tasks, researchers use the multi-task framework to carry out ECE task~\cite{chenJointLearningEmotion2018,wu2020multi}.
In addition, researchers take causal reasoning as an auxiliary task ~\cite{fan2020transition,turcanMultiTaskLearningAdapted2021a} to enhance the reasoning ability of the model.
However, most of these methods only focus on the interaction between tasks, without considering the dependency between clauses or utterances,
which is easy to cause long-distance information loss.
\citet{chen2020end,huFSSGCNGraphConvolutional2021} use GCN and other methods to capture the dependency between clauses.
Although these methods of ECE have realized the importance of interaction information between clauses,
they only use simple structural information such as distance to model the dependency between clauses.
We use discourse structures to make full use of the interaction information between utterances, which can be beneficial to encode long-distance dependency.

Most of the existing ECEC works do not consider the influence of the discourse relation between utterances and conversation-specific features.
Intuitively, utterance interactions information such as discourse structures are promising for the ECEC task.
In this paper, we employ discourse parsing as an auxiliary task by multi-task learning framework~\cite{he2021multi,fan2021multi} to capture utterance interaction information.
To model conversation-specific features, we refer \citet{wang2021structure,gatedgnn2016} to utilize a gated graph neural network~(gated GNN).
We also use gated GNN to further enhance discourse structure by encoding a discourse graph.
Finally, a gate controlled mechanism is applied to alleviate the error propagation problem from predicted discourse structures.

\section{Our Proposed Model}\label{sec:methodlogy}

\begin{figure*}
\centering
\includegraphics[width=1.8\columnwidth]{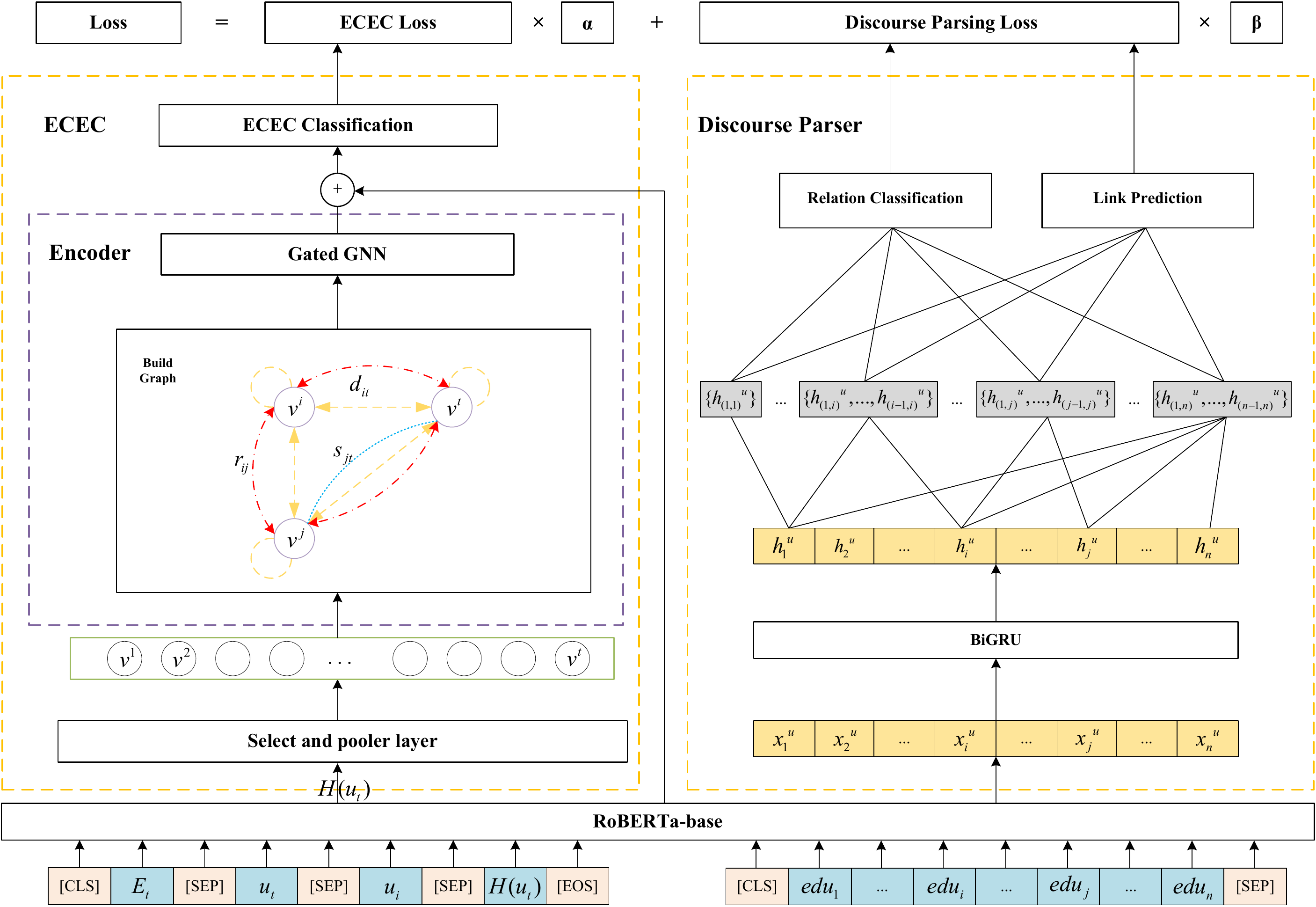}
\caption{Framework of our model.} 
\label{fig2}
\end{figure*}
In this section, we present the proposed discourse-aware model~(DAM) for ECEC.
As shown in Fig.~\ref{fig2}, 
we use an MLT framework to model the ECEC task and discourse parsing jointly. 
It can integrate rich utterance interaction information to our conversational context representations.
We use hard parameter sharing to combine these two task, and use RoBERTa~\cite{liu2019roberta} to obtain the shared conversational context representations.

To further enhance utterance interaction information, we encode conversational discourse structures by gated graph neural network~(gated GNN).
In addition, we use gated GNN to encode conversation-specific features such as speakers and relative utterance distance.
Concretely, we use the gate controlled mechanism to alleviate the error propagation problem caused by predicted discourse relations.
After that, we concatenate them with global encoding information and through a linear layer to obtain the classification result of the ECEC task.

As for auxiliary task discourse parsing, 
we use GRU to further obtain high-level hidden representations~(BiGRU)~\cite{cho2014learning}. 
Then we feed these high-level hidden representations to link predictor and relation classifier to obtain the corresponding discourse relations.

\subsection{Multi-Task Learning Framework}

\subsubsection{ECEC Model}
Given a conversation with $n$ utterances $\{u_1, u_2,..., u_n\}$, 
for each target utterance $u_t$ with emotion $E_t$ and its historical utterances set $H(u_t)$, 
the ECEC task aims to extract the cause utterances set $C(u_t)$ of the emotion expressed by $u_t$.
Therefore, this task can be regarded as a triple classification~\cite{poria2021recognizing}.
For each utterance $u_i$ in $H(u_t)$, the triple $(u_t, u_i, H(u_t))$ is taken as a positive case if $u_i \in C(u_t)$.
Otherwise, it is classified as a negative case.

\paragraph{Utterance Representations}
To compare with the recent Transformer-based model~\cite{poria2021recognizing}, we also use a [\texttt{CLS}] token and the emotional label [\texttt{E$_t$}] of the target utterance in front.
We concatenate the elements of triplet as extra inputs, as shown in Fig.~\ref{fig2}. 
Before building the graph of gated GNN, we preprocess the input of the ECEC task at the select and pooler layer.
We extract the hidden states $\mathbf{v}_{cls}$ of the [\texttt{CLS}] token as well as the hidden states of all tokens belonging to $H(u_t)$ for later use. 
The hidden states of tokens belonging to the DAMe utterance are represented as: $\{\mathbf{v}^{i}_{1},\mathbf{v}^{i}_{2},...,\mathbf{v}^{i}_{m_i}\}$,
where $i$ represents the $i^{th}$ utterance in $H(u_t)$, and $m_i$ represents the $i^{th}$ utterance with total of $m$ tokens.
For convenience and simplicity, these hidden states are summed as the utterance level representation: $\mathbf{v}^i=\sum\limits_{j=0}^{m_i} \mathbf{v}^{i}_{j}$.

\paragraph{ECEC Encoder}
In the ECEC Encoder, we build a full connected graph according to the discourse relations, speaker, and relative utterance distance.
Then, we use the gated GNN layer to encode inter-utterance representations and outputs hidden states $\mathbf{h}_i$.
See more details in the Section~\ref{subsec:ECEC}.

\paragraph{ECEC Classification}
After getting the output hidden states $\mathbf{h}_i$ of the encoder, 
we now use softmax function to get the probability distribution $\mathbf{P}_i$ through a linear layer and generate the prediction $\widehat{y}_i$ of inputs as follows:
\begin{equation}
\mathbf{P}_i=\mathrm{softmax}(\mathbf{W}_s \mathbf{h}_i+\mathbf{b}_s)
\end{equation}
\begin{equation}
\widehat{y}_i=\mathop{\arg\max}(\mathbf{P}_i)
\end{equation}
where $\mathbf{W}_s, \mathbf{b}_s$ are the weight matrix and bias of the linear before softmax.

\paragraph{Loss Function}
To train the model, we calculate the negative log-likelihood of train data as loss of the main task $L_e$ as follows:
\begin{equation}
L_e= - \sum_{v \in y_V} \sum_{z=1}^Z \mathbf{Y}_{vz} \log \mathbf{P}_{vz}
\end{equation}
where $y_V$ is the gold labels set and $Y$ is the label indicator matrix.

\subsubsection{Discourse Parser}

Given a conversation with $n$ elementary discourse units~(EDU) $\{edu_1, edu_2, ..., edu_n\}$, 
discourse parsing aims to predict dependency links and the corresponding relation types $\{(edu_j, edu_i, r_{ji})|j \neq i\}$ between the EDUs, where $(edu_j, edu_i, r_{ji})$ stands for a link of relation type $r_{ji}$ from $edu_j$ to $edu_i$.

\paragraph{Discourse Representations}
All EDUs from a conversation are fed into RoBERTa$_{\textrm{BASE}}$ to encode token-level information.
For $i^{th}$ EDU $edu_i$, the [\texttt{CLS}] token embedding is taken as the representation of $edu_i$, denoted as $\mathbf{x}_i^u$.
We use Bi-directional GRU~(BiGRU) to encode clause-level contextual information.
Specifically, we take the representation of EDUs $\{\mathbf{x}_1^u, \mathbf{x}_2^u, ..., \mathbf{x}_n^u\}$ as the input of BiGRU to get hidden representation $\{\mathbf{h}_1^u, \mathbf{h}_2^u, ..., \mathbf{h}_n^u\}$.

Then we follow~\citet{shi_deep_2019}, taking $\mathbf{h}_{<i,i} = \{\mathbf{h}_{(1,i)}, ..., \mathbf{h}_{i-1,i}\}$ as input of link predictor and relation classifier.
The Link predictor predicts the parent of $edu_i$.
Relation classifier is responsible for predicting the relationship type $r_{ji}$ between $edu_i$ and $edu_j$, if predicted parent of $edu_i$ and $edu_j$.

\paragraph{Link Prediction and Relation Classification}
Link predictor and relation classifier have similar structures.
They first converts the input vector $\mathbf{h}_{i,j}(j<i)$ into a hidden representation through a linear layer:
\begin{equation}
    \mathbf{L}_{i,j}^\mathcal{D}  = \mathrm{tanh}(\mathbf{W}_\mathcal{D}\mathbf{h}_{i,j} + \mathbf{b}_\mathcal{D})
\end{equation}
where $\mathbf{W}$ and $\mathbf{b}$ are learnable weights,
and $\mathcal{D} \in \{link, rel\}$ represent the weights of link predictor and relation classifier respectively.
Link predictor adopts softmax function to obtain the probability that $edu_j$ is the parent $p_i$ of $edu_i$ as follows:
\begin{equation}
    \mathbf{P}(p_i) = \mathrm{softmax}(\mathbf{W'}_{link}\mathbf{L}_{i,j}^{link} + \mathbf{b'}_{link})
\end{equation}
where $\mathbf{W'}_{link} \in \mathbf{R}^{1 \times d_l}$ and $\mathbf{b'}_{link} \in \mathbf{R}$ also is learnable weights, and $d_l$ is the dimension of $\mathbf{L}_{i,j}^{link}$.
Hence, the predicted $p_i$ is chosen as follows:
\begin{equation}
    p_i = \mathop{\arg\max}(\mathbf{P}(p_i))
\end{equation}
The relation classifier predicts the relation type $r_{ji}$ as follows:
\begin{equation}
    \mathbf{P}(r) = \mathrm{softmax}(\mathbf{W'}_{rel}\mathbf{L}_{i,j}^{rel} + \mathbf{b'}_{rel})
\end{equation}
$\mathbf{W'}_{rel} \in \mathbf{R}^{K \times d_r}$ and $\mathbf{b'}_{rel} \in \mathbf{R}^K$ are learnable weights. 
$K$ is number of relation types and $d_r$ is the dimension of $\mathbf{L}_{i,j}^{rel}$.

\paragraph{Loss Function}
As DAMe as ECEC, we calculate the negative log-likelihood loss as follows:
\begin{equation}
    L_{link} = - \sum_{d} \sum_{i=1}^n \log \mathbf{P}(p_i = p^*_i)
\end{equation}
\begin{equation}
    L_{rel} = - \sum_{d} \sum_{i=1}^n \log \mathbf{P}(r_{ji} = r^*_{ji})
\end{equation}
\begin{equation}
    L_{dp} = L_{link} + L_{rel}
\end{equation}
where $d$ is a conversation of train data and $p^*_i$ and $r^*_{ji}$ is golden labels.

\subsubsection{Training}
During training, two models are learned simultaneously for the two sub-tasks (ECEC and discourse parsing).
The bottom RoBERTa encoding layer is shared by the above two tasks.
The total training objective is defined as:
\begin{equation}
L=\alpha L_e +\beta L_{dp}
\end{equation}
where $\alpha$ and $\beta$ are hyperparameters.

\subsection{ECEC Encoder}\label{subsec:ECEC}

\subsubsection{Graph Construction}
Following \citet{wang2021structure,gatedgnn2016}, we adopt a gated graph neural network~(gated GNN) to capture conversation-specific features.
A fully connected graph is the input of the gated GNN, and each utterance in $H(u_t)$ is regarded as a node of the graph.
We explain the construction rules of discourse relations, speaker, and relative distance graph separately as follows.

\paragraph{Discourse Relation Graph:}
In order to model discourse relations between two utterances in conversations, 
we refer to the discourse parser proposed by \citet{wang2021structure} to obtain the discourse relations between utterances. 
We train discourse parser by the STAC dataset~\cite{afantenosDiscourseParsingMultiparty2015}.
Then, we input an entire conversation into the discourse parser:
\begin{equation}
\{(i,j,r_{ij}),...\}= \texttt{Parser}(u_1,...,u_n)
\end{equation}
where $(i,j,r_{ij})$ denotes a link of relation type $r_{ij}$ from $u_i$ to $u_j$, noting that $i,j=1,2,...,n$ and $j>i$.
We connect an edge if there is a discourse relation between any two nodes.
We give different weights to different discourse relation types.
These predicted discourse relations are not exactly correct, so it may lead to error propagation in the later use.

\paragraph{Speaker Graph:} 
Plain text is a continuous document written by a single author, while conversation is the interaction between multiple speakers, which is the most obvious difference between plain text and conversation.
Therefore, we incorporate conversation-specific features to the model by a fully connected graph.
A DAMe type of edge connects two adjacent utterances of the DAMe speaker.

\paragraph{Relative Distance Graph:} 
For the ECEC task, it is very important to establish the long-distance dependency between utterances.
In order to better model this dependency, we add relative utterance distance features.
There is an edge between any two nodes and a self-loop edge on each node. 
It represents the relative utterance distance between its linked nodes.

\subsubsection{Gated GNN}
Each node and edge in the graph represents as a learnable vector for gated GNN.
We directly use the utterance level representation $\mathbf{v}^i$ to initialize the vector representation of the corresponding node.
As for the edge between any two nodes, we initialize it by a learnable vector $\mathbf{e'}_{ij}$ as follows.
\begin{equation}
\mathbf{e'}_{ij}=[\mathbf{s}_{ij}, \mathbf{d}_{ij}, \mathbf{r}_{ij}]
\end{equation}
where $\mathbf{s}_{ij}$, $\mathbf{d}_{ij}$, $\mathbf{r}_{ij}$ respectively denote the speaker, relative utterance distance, and discourse relation type vectors between utterance $u_i$ and utterance $u_j$. 
It should be noted that the discourse relation between utterances is predicted by the discourse parser trained on STAC datasets.
Therefore, there is a problem of error propagation.

After direct concatenation of three vectors, we adopt a learnable gate module in $\mathbf{e'}_{ij}$ to selectively forget certain information in neural networks.
It can control duplicated discourse information as well as relieving the issue of error propagation, according to our experiments. 
\begin{equation}
    \mathbf{e}_{ij}=\mathbf{e'}_{ij}\odot\sigma(\mathbf{W}_f\mathbf{e}_{ij}'+\mathbf{b}_f)
\end{equation}
where $\sigma$ and $\odot$ represents $\rm{sigmoid}$ function, dot-product operation separately, $\mathbf{W}_f,\mathbf{b}_f$ are weight matrix and bias.

After that, we conduct structure-aware scaled dot-product attention to update the hidden state of nodes.
More details can be seen in \citet{wang2021structure}. 
We will iterate $T$ times to update the above hidden states.
Eventually, we concatenate the hidden state vectors of two directions $\mathbf{e}_{ij}^{T}, \mathbf{e}_{ji}^{T}$ in the top layer:
\begin{equation}
\mathbf{\widehat{E}}_{i,j}=[\mathbf{e}_{ij}^{T};\mathbf{e}_{ji}^{T}]
\end{equation}
where the subscript $i,j$ represents the $i{th}$ and $j^{th}$ utterance.
The GNN encoding representation $\widehat{E}_{i,t}$ and the hidden representation $\mathbf{v}_{cls}$ of [\texttt{CLS}] in previous stage are concatenated together as the final representation $\mathbf{h}_i$:
\begin{equation}
\mathbf{h}_i=[\mathbf{v}_{cls};\mathbf{\widehat{E}}_{i,t}]
\end{equation}
where $i$ represents the $i^{th}$ utterance that need to be classified and $t$ represents the target utterance that contains the non-neural emotion.

\section{Experiment Settings}\label{sec:experiments}
\subsection{Datasets}
To verify the effectiveness of utterance interactions information, we conduct experiments on the RECCON~\cite{poria2021recognizing} dataset. 
We use the Fold1 dataset \cite{poria2021recognizing} generated by a negative DAMpling strategy as our experimental dataset. 
Training, validation, and testing data are 27915, 1185, and 7224 DAMples respectively.

Moreover, we use the STAC~\cite{afantenosDiscourseParsingMultiparty2015} dataset to help complete the training of auxiliary task discourse parsing. 
There are 1091 conversations with 10677 EDUs and 11348 discourse relations in STAC.

\subsection{Hyperparameters}
We use RoBERTa$_{\textrm{BASE}}$ to encode the conversational contexts, and adopt an AadmW algorithm to optimize the parameters of our model.
The initial learning rate is set to 1e-5, batch size sets to 8.
$\alpha, \beta$ set to 1 and 0.25.
The dimensions for the edge and the node states in gated GNN are both 768.
And the dimensions for $\mathbf{s}_{ij}, \mathbf{d}_{ij}$ and $\mathbf{r}_{ij}$ are set to be 192, 192 and 384, respectively. 
We train 10 epochs on the training set and save the best model according to the performance on the validation set.
Based on the best model, we test the performance on the test set.

\subsection{Evaluation}
For fair comparison, we use F1-scores to evaluate our proposed model.
Pos.F1 and Neg.F1 represent the F1-score on the positive and the negative examples, respectively.
In addition, we report the overall MacroF1 score based on the Pos.F1 and the Neg.F1.

\subsection{Benchmark Models}
\begin{table}
    \centering
    \begin{tabular}{lccc}
            \toprule
            \textbf{Model}         & \textbf{Pos.F1}    & \textbf{Neg.F1}   & \textbf{MacroF1}  \\
            \midrule
            RankCP        & 33.00     & 97.30    & 65.15     \\
            ECPE-MLL      & 48.48     & 94.68    & 71.59     \\
            ECPE-2D       & 55.50     & 94.96    & 75.23     \\
            SIMP$_{\textrm{BASE}}$  & 64.28     & 88.74    & 76.51    \\
            SIMP$_{\textrm{LARGE}}$ & 66.23     & 87.89    & 77.06    \\
            \midrule
            DAM~(ours)   & {\bfseries 67.91}     & 89.55    & {\bfseries 78.73}     \\
            \bottomrule
    \end{tabular}
     \caption{The results of our model in comparison to other models, the SIMP$_{\textrm{BASE}}$ and SIMP$_{\textrm{LARGE}}$ are Transformer-based models proposed by \citet{poria2021recognizing} initializing with RoBERTa$_{\textrm{BASE}}$ and RoBERTa$_{\textrm{LARGE}}$ respectively.}
    \label{tab1}
\end{table}
We first compare our DAM with three ECE SOTA models proposed by previous research in Table~\ref{tab1}.
\begin{itemize}
\item[1.] RankCP~\cite{wei2020effective}: It used graph attention network to encode the representation of document clause;
\item[2.] ECPE-2D~\cite{ding2020ecpe}: It integrated the representation, interaction and prediction of 2D emotion-cause pairs by jointing learning;
\item[3.] ECPE-MLL~\cite{ding2020end}: It used the multi-label learning scheme for training, and obtained a good effect.
\end{itemize}
\citet{poria2021recognizing} transfer these three models to conversation dataset and propose some simple but better Transformer-based models (SIMP$_{\textrm{BASE}}$ and SIMP$_{\textrm{LARGE}}$). 

\subsection{Main Results}
We can draw the following conclusions. 
First, because of the differences between plain text and conversation, the previous complex deep learning methods such as RankCP, ECPE-MLL and ECPE-2D for plain text document data perform even worse than the simple SIMP$_{\textrm{BASE}}$ model.
Second, our model improves by nearly 2.22\% over the baseline model encoded by the DAMe RoBERTa$_{\textrm{BASE}}$.
This can be owing to the potential links between emotional utterances and their corresponding cause expressions contained in the discourse structures. 
Conversation-specific features can also help models understand conversations.

\section{Results and Analysis}\label{sec:results}
We next conduct more experiments to explore which part of our model works and what we have promoted.
We also study how to combine these structure information so that making the most of them.

\subsection{Ablation Study}

\begin{table}
    \centering
    \resizebox{0.48\textwidth}{!}{
    \begin{tabular}{lccc}
            \toprule
            \textbf{Model} & \textbf{Pos.F1}    & \textbf{Neg.F1}   & \textbf{MacroF1}   \\
            \midrule
            DAM                  & 67.91     & 89.55    & 78.73     \\
            \midrule
            W/O multi-task         & 65.76     & 87.44    & 76.60     \\
            W/O gated-gnn            & 66.61     & 88.13   & 77.37     \\
            \midrule
            W/O speaker            & 65.32     & 89.38    & 77.35     \\
            W/O distance           & 66.58     & 88.44    & 77.51     \\
            W/O gate        & 66.32     & 89.02    & 77.67     \\
            \bottomrule
    \end{tabular}
    }
    \caption{\label{tab2}Ablation results. ``W/O'' represents ``without"}
\end{table}
Previous main results show that the effectiveness of our model.
As shown in Table~\ref{tab2}, the performance of removing auxiliary task decreases by 2.13\% over the DAM.
It indicates that multi-task framework can help our model capture discourse structures information.
This information improves our model's ability to establish relationships between utterances.
The second row shows the model without gated GNN decreases by 1.36\% over the DAM, showing the help of encoding conversation context information for understanding conversation.
We also verify the influence of speakers and relative utterance distance features respectively, the performance of them all decrease.
It demonstrates that these structure information containing conversational features have a significant impact on the model. 
Finally, it can be found that only remove the gate mechanism, performance decreases by 1.06\%.
We argue that the existence of the gate module alleviates the error propagation caused by directly using the pseudo discourse relations.

\subsection{Long-distance Dependency Analysis}

\begin{figure}[t]
    \centering
    \includegraphics[width=0.85\columnwidth]{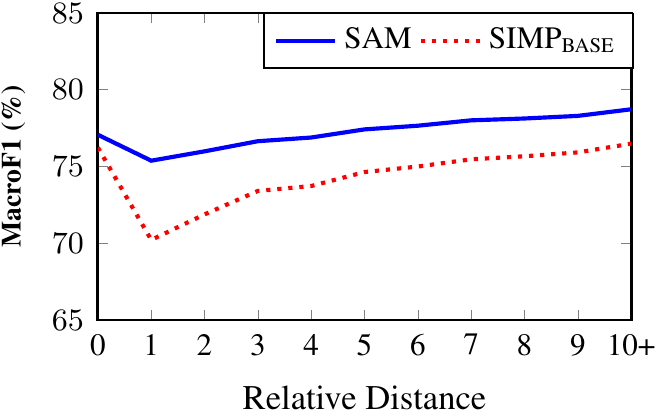}
    \caption{Performance of different relative utterance distance between $u_t$ and $u_i$.} 
    \label{fig3}
\end{figure}

To verify whether our model can alleviate the long-distance dependency problem, we present a set of results of different relative utterance distance between $u_t$ and $u_i$ in Fig~\ref{fig3}.
Foremost, we can find whatever the relative utterance distance are, our DAM model performs better than previous SIMP$_{\textrm{BASE}}$ model.
That means our model is more robust in terms of distance by combining the utterance interaction features.
Then we see that our model performs significantly better than the SIMP$_{\textrm{BASE}}$ inside a turn or distance of 3.
However, above 3, the performance does not significantly increase.
It suggests that in the ECEC task, long-distance dependency problem is significant.
Although our approach can help to ameliorate the problem, deep reasoning needs to be further explored in the case of overextended turns or distance.

\subsection{Effect of Different Fusion Methods}
\begin{figure*}
    \centering
    \includegraphics[width=1.9\columnwidth]{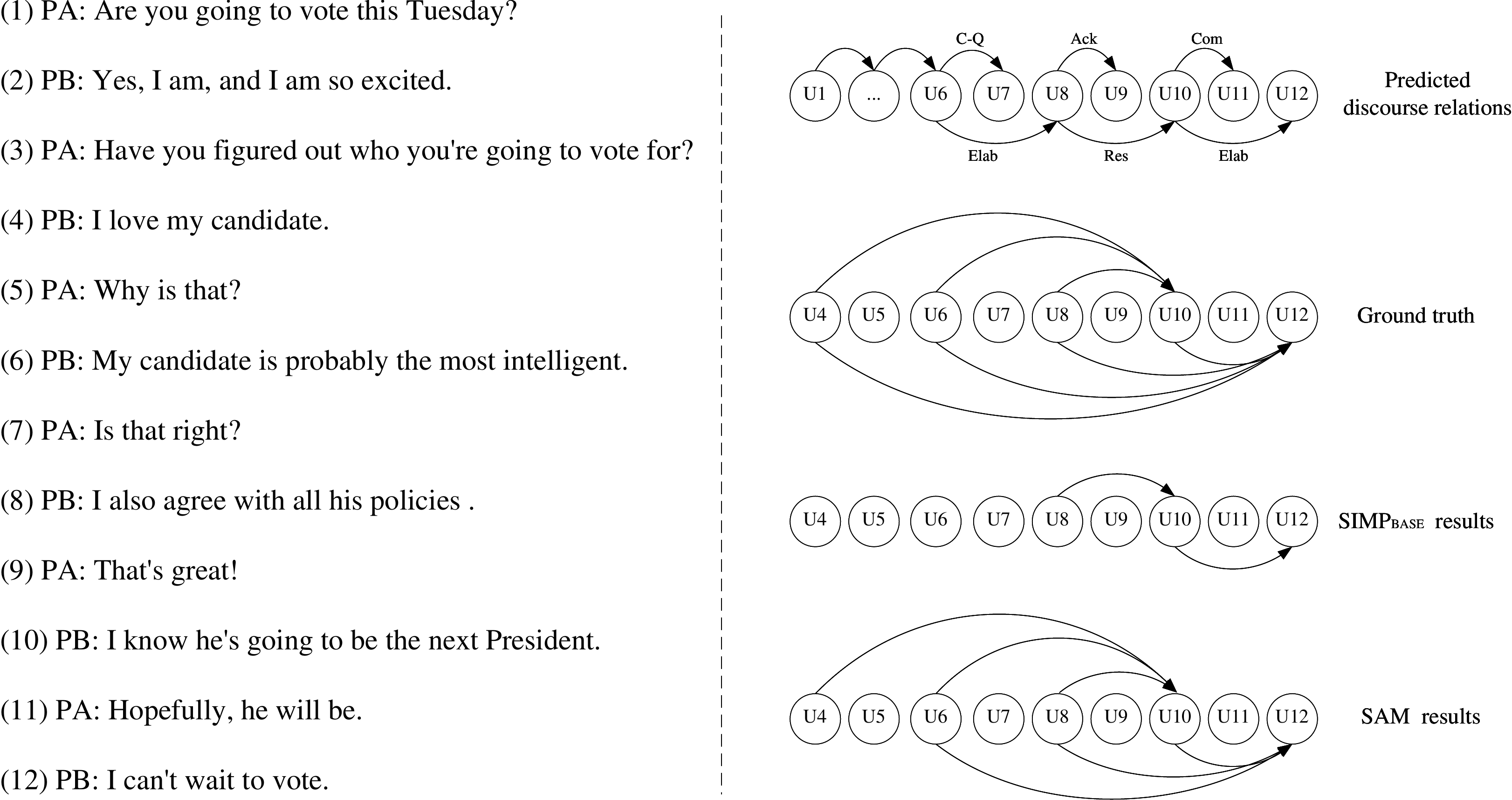}
    \caption{A conversation example of test dataset along with predicted discourse relations, Ground truth, SIMP$_\textrm{BASE}$ results and DAM results. 
    "C-Q" is short for "Clarification\_question", "Elab" for "Elaboration", "Res" for "Result", "Ack" for "Acknowledgement", and "Com" for "Comment". 
    $u_i$ corresponds to the $i_{th}$ utterance of the conversation.} 
    \label{fig4}
\end{figure*}

\begin{table}
\centering
\begin{tabular}{lccc}
\toprule
\textbf{Model} & \textbf{Pos.F1}    & \textbf{Neg.F1}   & \textbf{MacroF1}  \\
\midrule
SIMP$_{\textrm{BASE}}$  & 64.28     & 88.74    & 76.51     \\
SIMP$_{\textrm{LARGE}}$ & 66.23     & 87.89    & 77.06     \\
\midrule
DAM          & 67.91     & 89.55    & 78.73     \\
DAM-mtl      & 66.61     & 88.13    & 77.37     \\
DAM-cat      & 65.04     & 88.36    & 76.70     \\
DAM-gnn      & 66.91     & 89.24    & 78.07     \\
DAM-mtl-gnn  & 66.32     & 89.02    & 77.67     \\
DAM-arc      & 66.70     & 88.87    & 77.79     \\
\bottomrule
\end{tabular}
\caption{\label{tab5}Result of different fusion methods of discourse relations. DAM is our final strategy.}
\end{table}

In this section, we study how to integrate discourse relations so that making the most of them.
\citet{chen2020end,ding2020ecpe} use graph convolutional network or joint learning to model the dependency relations between clauses.
In this paper, we design and experiment with six alternative methods for incorporating discourse relations.
Table~\ref{tab5} show the performance of these methods, and we can draw the following conclusions.
First, we can see that discourse structure information can be strengthened by three strategies (DAM-mtl/cat/gnn).
Therefore, bringing better performance than the SIMP$_{\textrm{BASE}}$ that initialized by RoBERTa$_{\textrm{BASE}}$.
However, the improvement of DAM-cat is slight.
It may due to that it suffers from the serious error propagation directly using not exactly correct hidden states.
When we combine the DAM-mtl and DAM-gnn~(DAM-mtl-gnn), the performance degrades.
It may be due to the duplicated discourse information or error propagation.
We add the gate module (DAM) on the basis of DAM-mtl-gnn that  effectively alleviates the above problems and performs best.
So we choose this fusion method as final method.
Moreover, we also try to not use the specific discourse relation types and instead focus on whether they have discourse relations~(DAM-arc) or not, but it still does not comparable with DAM.

\subsection{Case Study}

For the case study, we select an example in the test dataset to verify that discourse structure information can help our model to better model long-distance dependencies.

Due to space limitations, we only show a subset of emotional cause and discourse relations.
As shown in Fig.~\ref{fig4}, the cause utterances set of $u_{10}$ is $C(u_{10}) = {u_4, u_6, u_8}$, and the cause utterances set of $u_{12}$ is $C(u_{12}) = {u_4, u_6, u_8, u_{10}}$.
The SIMP$_{\textrm{BASE}}$ model can only predict the cause utterances with a short relative distance, and its performance ability is not good for the cause utterances with a long relative distance.
It can be seen from the discourse relationship diagram that the target utterances and each utterance in the cause utterances set are related to each other.
After we incorporate the discourse structure information into the model, the model can better capture long-distance dependencies.
Although there are still some cause utterances that have not been successfully found, such as the cause utterance $u_4$ of the target utterance $u_{12}$, we guess that this is caused by the error propagation of the discourse parser.

\section{Conclusion and Future Work}\label{sec:conclusion}
In this paper, we introduce a discourse-aware model for emotion cause extraction in conversation.
Specifically, we model ECEC with discourse parsing in conversations by multi-task framework.
It can help share pre-trained language model learning better discourse structure representations of conversations. 
In addition, this model employ a graph neural network to encode conversation-specific features such as relative utterance distance, and speakers, and further enhance discourse structures.
Both of them can help the model integrate rich utterance interactions information and mitigate long-distance dependency problem.
Finally, we utilize a gate controlled module to alleviate the error propagation problem from predicted discourse relations.
Experiments on the benchmark show that our model reaches the new SOTA.

Our future works may explore how discourse structure information can be used to extract both the emotion and the cause without the target utterance emotion information.
Besides, we still need to explore more methods to solve the problem of long-distance dependence and error propagation.

\bibliography{custom}
\bibliographystyle{acl_natbib}

\end{document}